\documentclass[runningheads]{llncs}
\usepackage{graphicx}

\usepackage{tikz}
\usepackage{comment}
\usepackage{amsmath,amssymb} % define this before the line numbering.
\usepackage{color}
\usepackage{enumitem}
\usepackage{booktabs}
\usepackage{xcolor}
\usepackage{units}
\usepackage{bbding}
\newcommand{\changed}[1]{{\color{black}{#1}}}
\newcommand{\name}{Ego3RT}
\newcommand{\backbone}{Image feature extractor}
\newcommand{\neck}{Back tracing decoder}

\newcommand{\attention}{MVAA}
\usepackage{float}
\usepackage{subfig}
\usepackage[labelfont=bf]{caption}
\usepackage[pagebackref=true,breaklinks=true,colorlinks,bookmarks=false]{hyperref} 
\usepackage{colortbl}
\definecolor{maroon}{cmyk}{0,0.87,0.68,0.32}

\usepackage{cite}

\usepackage{multirow}
\usepackage{array}
\newcommand{\PreserveBackslash}[1]{\let\temp=\\#1\let\\=\temp}
\newcolumntype{C}[1]{>{\PreserveBackslash\centering}p{#1}}

\usepackage{lipsum}

\def\wrt{\textit{w.r.t}}
\def\eg{\textit{e.g.}}
\def\ie{\textit{i.e.}}

\begin{document}

\pagestyle{headings}
\mainmatter

\title{Learning Ego 3D Representation as Ray Tracing}

\titlerunning{Learning Ego 3D Representation as Ray Tracing}

\author{Jiachen Lu$^1$ \and
Zheyuan Zhou$^1$ \and
Xiatian Zhu$^2$ \and
Hang Xu$^3$ \and
Li Zhang$^1$\thanks{Li Zhang (lizhangfd@fudan.edu.cn) is the corresponding author with School of Data Science, Fudan University.
}
\vspace{-0.5em} 
}

\authorrunning{J. Lu, Z. Zhou, et al.}

\institute{
$^1$Fudan University \quad
$^2$University of Surrey \quad
$^3$Huawei Noah's Ark Lab 
\\
\vspace{0.5em} 
\url{https://fudan-zvg.github.io/Ego3RT}
}

\maketitle
\begin{figure}[h]
    \centering
    \vspace{-2em}
    \includegraphics[width=\linewidth]{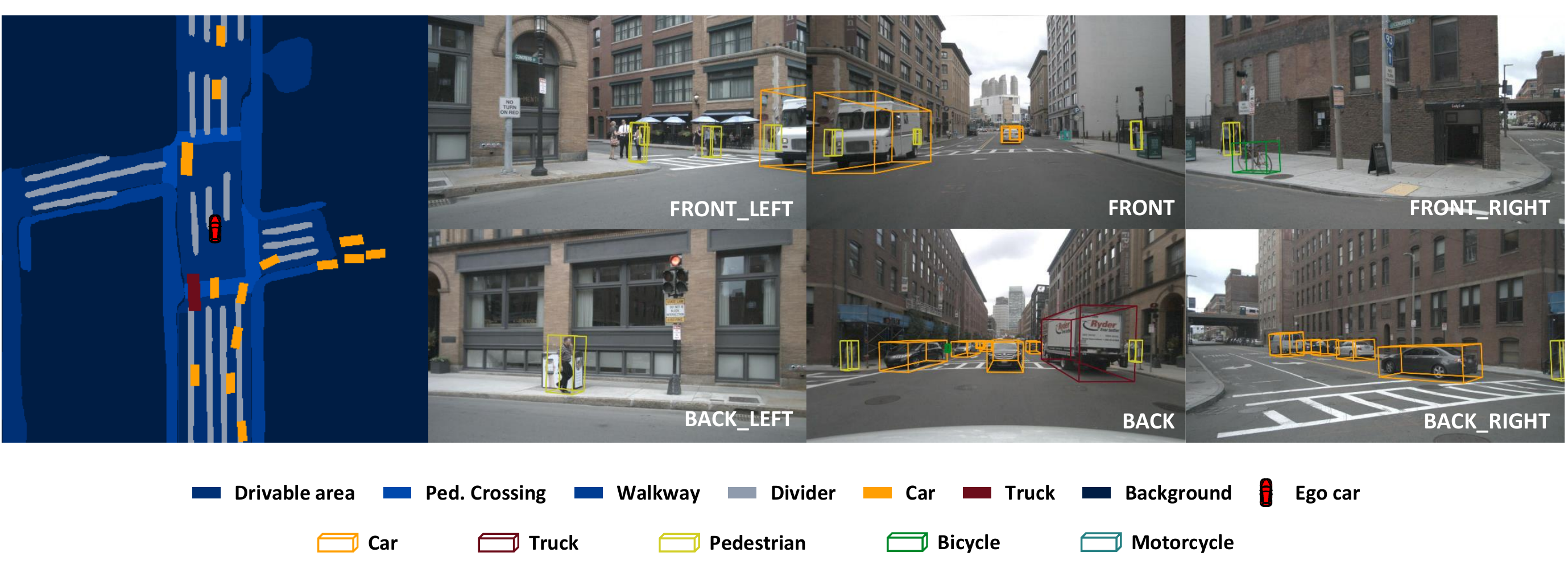}
    \vspace{-5mm}
    \caption{We propose an ego 3D representation learning method that extracts 3D representation in bird's-eye-view (BEV) 
    % coordinate frame 
    from multi-view cameras ({\em right}).
    With the aid of our 3D representation, multiple tasks can be executed efficiently in a single model: BEV segmentation ({\em left}) and 3D object detection ({\em right}).
    }
    \vspace{-8mm}
    \label{fig:intro-fig}
\end{figure}
\begin{abstract}
A self-driving perception model aims to extract 3D semantic representations from multiple cameras collectively into the bird's-eye-view (BEV) coordinate frame of the ego car in order to ground downstream planner.
Existing perception methods often rely on error-prone depth estimation of the whole scene or learning sparse virtual 3D representations without the target geometry structure, both of which remain limited in performance and/or capability.
In this paper, we present a novel end-to-end architecture for ego 3D representation learning from an arbitrary number of unconstrained camera views.
Inspired by the ray tracing principle, we design a polarized grid of ``imaginary eyes" as the learnable ego 3D representation and formulate the learning process with the adaptive attention mechanism in conjunction with the 3D-to-2D projection.
Critically, this formulation allows extracting rich 3D representation from 2D images without any depth supervision, and with the built-in geometry structure consistent \wrt~BEV. 
Despite its simplicity and versatility, extensive experiments on standard BEV visual tasks (\eg, camera-based 3D object detection and BEV segmentation) show that our model outperforms all state-of-the-art alternatives significantly, with an extra advantage in computational efficiency from multi-task learning.

\end{abstract}
\section{Introduction}

Taking an image as input, existing vision models
usually either ignore (\eg, image classification \cite{simonyan2014very, he2016deep, dosovitskiy2020image}) or consume directly  (\eg, object detection \cite{ren2015faster,misra2021end, zhu2020deformable}, image segmentation \cite{long2015fully, chen2017rethinking, zheng2021rethinking}) the coordinate frame of input during results prediction.
Nonetheless, this paradigm does not match the perception circumstance of self-driving out-of-the-box, where the input source is multiple cameras each with a specific coordinate frame, and the perception models for downstream tasks (\eg, 3D object detection, lane segmentation) need to make predictions in the coordinate frame of the ego car, totally {\em different} from all the input frames.
That is, a self-driving perception model needs to reason about 3D semantics from 2D visual representations of multi-view images,
which is non-trivial and highly challenging.

\begin{figure}[tb]
    \centering
    \includegraphics[width=\linewidth]{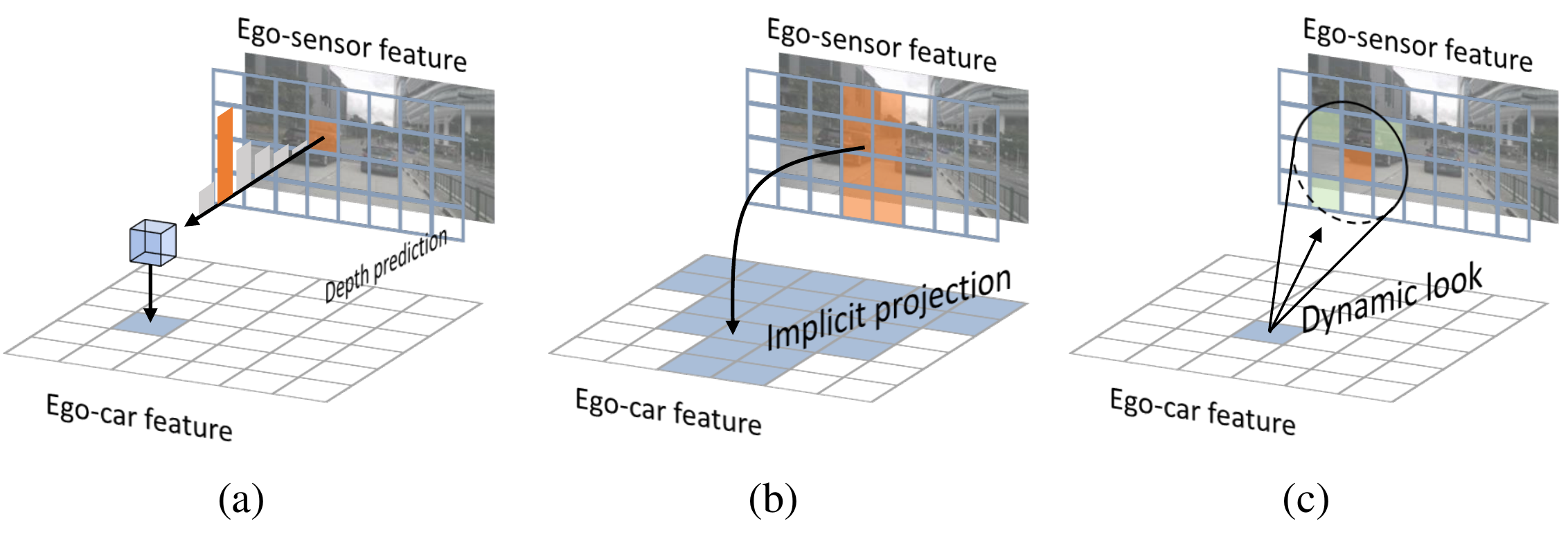}
    \caption{Comparison of dense 3D representation learning strategies. (a) The \textit{\textbf{first}} strategy, represented by LSS~\cite{philion2020lift}, CaDDN~\cite{reading2021categorical}, is based on dense pixel-level depth estimation. 
    (b) The \textit{\textbf{second}} strategy represented by PON~\cite{roddick2020predicting} bypasses the depth estimation by learning implicit 2D-3D projection.
    (c) {\bf\em Our} strategy that backtracks 2D information from ``imaginary" eyes specially designed in the BEV's geometry.
    }
    \label{fig:pipeline_comp}
\end{figure}

In the literature, existing methods mostly take the following two strategies.
The {\bf \em first} strategy show in Figure~\ref{fig:pipeline_comp}(a) (\eg, 
LSS~\cite{philion2020lift}, CaDDN~\cite{reading2021categorical}) relies on pixel-level depth estimation, as it can be used to project the 2D visual representation to the ego coordinate frame alongside intrinsic and extrinsic projection. 
Often, the depth prediction is end-to-end learned within the model without supervision~\cite{philion2020lift}, or with extra 3D supervision~\cite{reading2021categorical}.
A downside of these methods is that depth estimation in unconstrained scenes is typically error-prone,
which would be further propagated down to the subsequent components.
This is also known as the {\em error propagation} problem,
largely inevitable for such pipelines.

To solve this above issue,
the {\bf \em second} strategy (\eg, Image2Map \cite{saha2021translating}, OFT \cite{roddick2018orthographic}, DETR3D \cite{wang2022detr3d})
eliminates the depth dimension via directly learning 
3D representations from 2D images through architecture innovation.
This approach has shown to be superior over depth-estimation based counterparts,
implying that learning 3D representation is a superior general strategy.
In particular, Image2Map \cite{saha2021translating} and PON~\cite{roddick2020predicting} leverage a Transformer or FC layer to learn the
projection from 2D image frame to the bird’s-eye-view (BEV) coordinate frame forwardly.
As is shown in Figure~\ref{fig:pipeline_comp}(b), However, their 3D representation is structurally inconsistent with 2D counterparts as no rigorous intrinsic and extrinsic projection can be leveraged,
\ie, no explicit one-one correspondence relation
across the coordinate frames, 
consequently resulting in sup-optimal solutions.
The recent state-of-the-art DETR3D \cite{wang2022detr3d} formulates a 3D representation learning model
with a Transformer model,
inspired by contemporary image based object detection models \cite{carion2020end}. 
However, its 3D representation is not only {\em sparse}, but {\em virtual}
in the sense of no geometry structure explicitly involved \wrt~the ego coordinate frame,
and is thus unable to conduct dense prediction tasks such as segmentation.

In this work, we present a novel ego 3D representation learning method that overcomes all the aforementioned limitations in an end-to-end formulation.
This is inspired by the ray tracing principle~\cite{whitted2005improved} in computer graphics, which simulates the light transport process from the sources to human eyes in graphic rendering. 
Rather than taking the optical sources as inception, ray tracing {\em backtracks} the optical paths from the imaginary eyes to the objects in an opposite way.
Analogously, we start with introducing a polarized grid of dense ``imaginary eyes" for BEV representation, with each eye naturally occupying a specific geometry location with the depth information involved.
As is shown in Figure~\ref{fig:pipeline_comp}(c), for learning 3D representation including height information intrinsically absent in BEV, we initialize each eye using a uniform value and leave the eyes to look backward surrounding 2D visual representations subject to the intrinsic and extrinsic 3D-to-2D projection. 
With the adaptive attention mechanism, eyes focus dynamically on 2D representations and directly learn to approximate missing height information in a data driven manner.
Critically, our architecture can be applied for both sparse (\eg, 3D object detection)
and dense (\eg, BEV semantic segmentation) prediction tasks.
We term our method 
{\em Ego 3D representation learning as Ray Tracing} ({\bf\em Ego3RT}).

We make the following {\bf contributions}:
{\bf(1)} We propose a novel ego 3D representation learning architecture, inspired by the ray tracing perspective. 
{\bf(2)} Without depth supervision, our method can learn 
geometrically structured and dense 3D representations from arbitrary camera rigs \wrt~the ego car coordinate frame,
subject to the intrinsic and extrinsic 3D-to-2D projection.
This is achieved by adapting the ray tracing concept, where we first introduce a polarized grid of ``imaginary eyes" as the learnable BEV representation,
and then trace them {\em backwards} to camera rigs by formulating the learning process of this 3D representation into an adaptive attention framework.
{\bf(3)} Expensive experiments on 3D object detection and lane segmentation self-driving tasks
validate the superiority of our method over state-of-the-art alternative methods, often by a large performance gap.
In particular, \name~enables multi-task learning
by representation sharing between object detection and BEV segmentation
whilst still yielding superior performance,
hence more computationally efficient and economically scalable for self-driving.
\section{Related work}
\paragraph{\bf Depth-based strategy}
Benefited from well-studied depth estimation, Pseudo-lidar~\cite{wang2019pseudo}, Pseudo-lidar++~\cite{you2019pseudo} and AM3D~\cite{ma2019accurate} separate the 3D representation learning into monocular depth estimation and 3D detection.
CaDDN~\cite{reading2021categorical} uses a supervised depth estimation network to accumulate a more precise position of each voxel from the front-view features.
These dual-step methods rely on extra depth estimation data and are not end-to-end trainable.
LSS~\cite{philion2020lift} and FIERY~\cite{hu2021fiery} achieve the front-view features lifting in an end-to-end manner with the depth distribution prediction to generate the intermediate 3D representations.
However, weakly-constrained depth estimation is error-prone, with the depth error propagated to limit the subsequent 3D localization.

\paragraph{\bf Depth-free strategy}
OFT~\cite{roddick2018orthographic} simply hypothesizes a uniform distribution over the depth, 
leading to poor performance in the 3D detection task.
Rather than predicting depth, \cite{yang2021projecting, chitta2021neat, roddick2020predicting, saha2021translating} opt to exploit the 3D-to-2D projection process.
PYVA~\cite{yang2021projecting} shows that the correspondence between 2D features and 3D features can be implicitly learned by cross attention.
NEAT~\cite{chitta2021neat} further proposes a variation of cross attention for the same purpose.
PON~\cite{roddick2020predicting} and Image2Map~\cite{saha2021translating} resort to a Transformer or FC layer to learn a correspondence between images and 3D features.
While decently estimating the 3D-to-2D relationship, a clear limitation is that these models ignore the
intrinsic one-one correspondence.
Recently, DETR3D~\cite{wang2022detr3d} learns sparse 3D representations with a  Transformer by using sparse queries to detect 3D objects.
However, this 3D representation has no geometry structure, making it incapable of performing dense prediction tasks. 
In this work, we present \name, a novel end-to-end trainable ego 3D representation learning architecture that solves all the above limitations.
Critically, it can learn 3D ego representation from unconstrained camera rigs
without any 3D or depth supervision, achieving superior performance on BEV visual tasks even in a more efficient multi-task model design.
\section{Method}
\begin{figure}[t]
    \centering
    \includegraphics[width=0.9\linewidth]{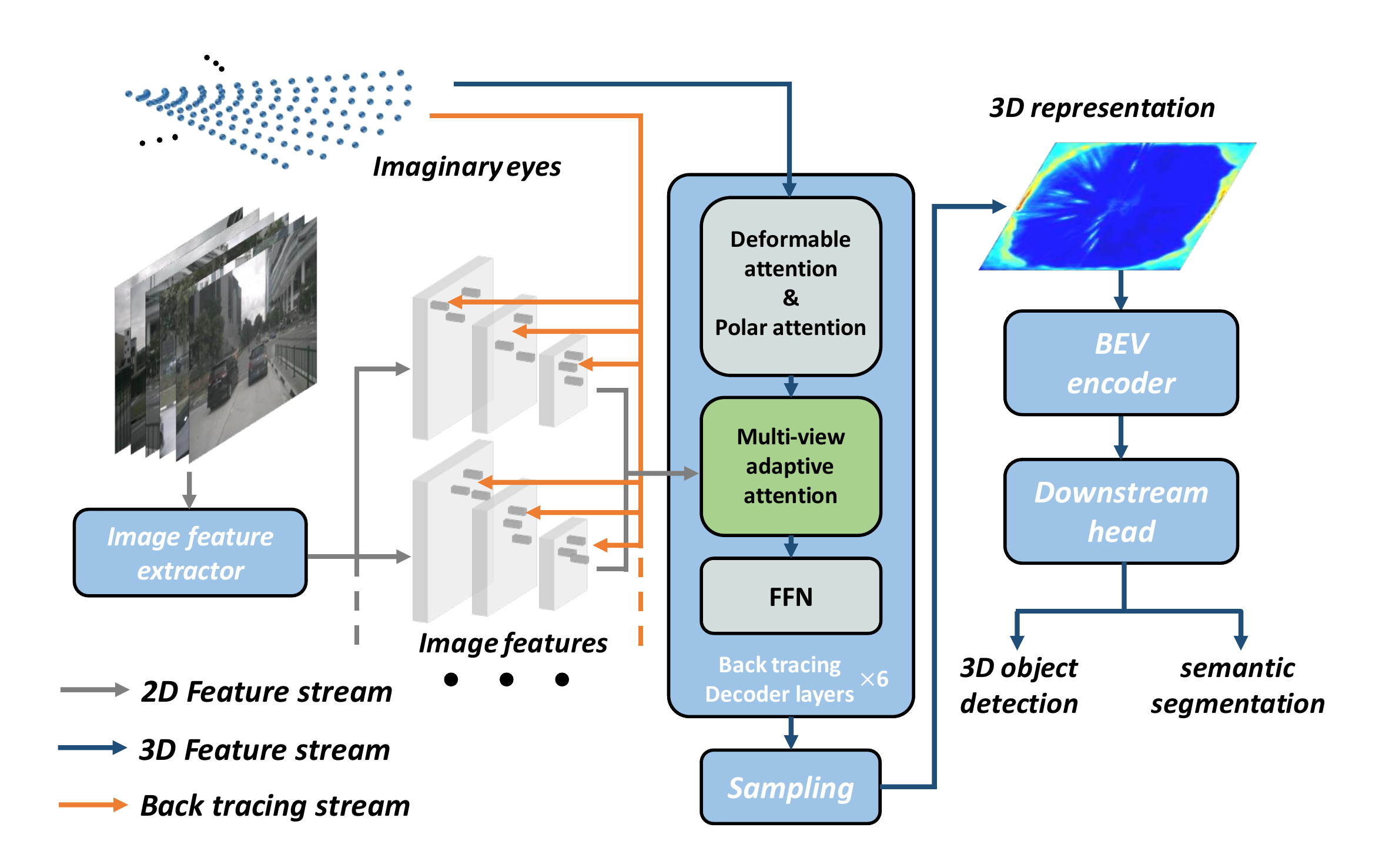}
    \caption{Our pipeline comprises two stages: learning ego 3D representation from 2D features and executing multiple downstream tasks based on 3D representation. The gray lines represent the 2D feature stream while the blue lines represent the 3D feature stream. Besides, the orange lines specify our back tracing path.}
    \label{fig:pipeline}
\end{figure}
Our architecture can be divided into two components: 
(1) ego 3D representation learning ({\bf\em\name}) and 
(2) downstream task head.

\subsection{\name: Ego 3D representation learning}
\name~consists of two parts: 
image feature extractor and back tracing decoder.
In addition, to illustrate back tracing decoder clearly, we will first introduce its components -- imaginary eyes, tracing 3D backwards to 2D mechanism and multi-view  multi-scale  adaptive  attention.
\paragraph{\bf \backbone}
Given a set of images $\mathcal{I} = \{ \mathbf{p}_1, \mathbf{p}_2, \cdots \mathbf{p}_{N_\text{view}} \}$ from multiple camera sensors (\ie, multiple views), where each image $\mathbf{p}_t\in \mathbb{R}^{H\times W\times 3}$ with $t$ the index of surrounding cameras (\eg, $N_\text{view}=6$ for nuScenes).
These images are then encoded by a single shared ego-sensor feature extractor, including a CNN and a transformer encoder.
ResNet~\cite{he2016deep} is used to extract image feature maps
at $N_\text{scale}$ spatial resolutions.
To capture global context information, we further apply a transformer encode \cite{zhu2020deformable} at each resolution individually.
As a result, 
we obtain the multi-view multi-scale self-attentive 2D representation
$\{\mathbf{x}_l^{(t)}\}_{l=1}^{N_\text{scale}}$, where $\mathbf{x}_l^{(t)}\in\mathbb{R}^{H_l\times W_l\times C_l}$, $H_l=\frac{H}{4*2^l}, W_l = \frac{W}{4*2^l}$, $t \in [1,\cdots,N_\text{view}]$.

\begin{figure}[t]
    \centering
    \includegraphics[width=0.7\linewidth]{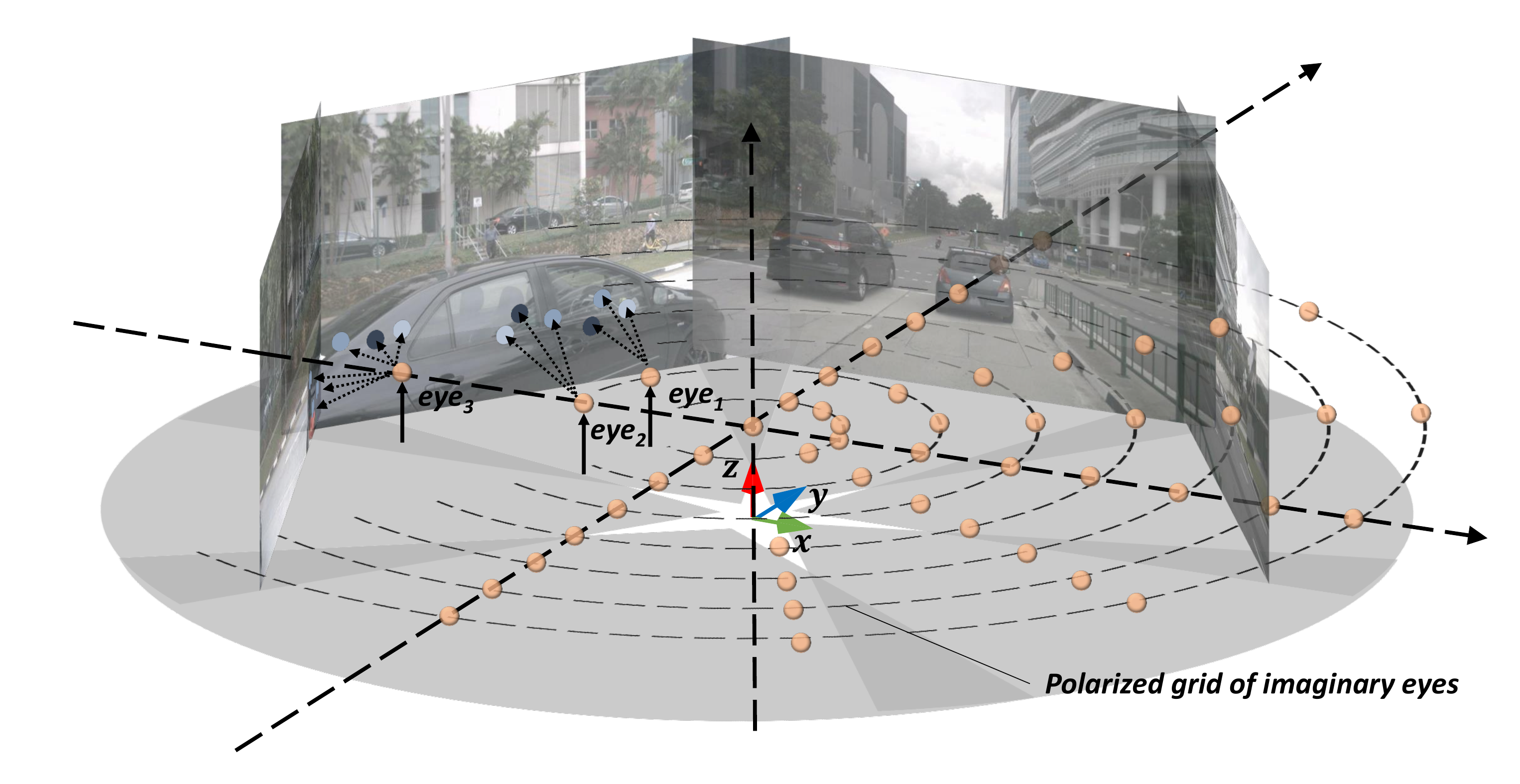}
    \caption{An illustration of tracing 3D backwards to 2D mechanism for imaginary eyes.
    The golden balls represents the polarized grid of dense ``imaginary eyes". 
    Specially, for eyes have multiple visible images (e.g. $\text{eye}_3$), they backtrack to multiple images, while eyes having only single visible image (e.g. $\text{eye}_1$) backtrack to single image.
    The blue points on image from light blue to deep show their degree of significance to their eye, thus facilitating the adaptive attention.
    }
    \label{fig:2d3dmapping}
\end{figure}

\paragraph{\bf Imaginary eyes}
From this part, we will specify how \name~learning 3D representation from 2D.
To avoid exhaustive pixel-level prediction and inconsistent coordinate projection, we draw an analogue in the back tracing idea from ray tracing~\cite{whitted2005improved}.
We start with introducing a polarized grid of dense ``imaginary eyes'' shown in Figure~\ref{fig:2d3dmapping}, for BEV representation, with each eye naturally occupying a specific geometry location with built-in depth information.
The grid of eyes are in size of $N_\text{eyes}= R \times S$, where $R$ is the number of eyes on each polar ray, and $S$ is the number of these polar rays.
To construct or, ``render" our BEV representation, 
these imaginary eyes send rays \textit{backwards} to 2D visual representation following the above 3D-2D projection routine (which will be described later).
Since each eye only occupies a single, fixed geometry location, tracing back at its corresponding 2D position alone is less informative due to limited local observation.
To solve this problem, we propose to encourage the eyes 
look around to focus adaptively on pivotal feature points across multiple scale per image and multiple camera views.
This results in 
our multi-view multi-scale adaptive attention module (\attention).
And finally, the features of these imaginary eyes will be the final 3D representation.

\paragraph{\bf Tracing 3D backwards to 2D}
To specify the tracing back mechanism, we first illustrate coordinate transformation between 3D and 2D.
In typical cases, we usually have one LIDAR coordinate (3D), $N_\text{view}$ camera coordinate (3D) and $N_\text{view}$ image coordinate (2D). 
First, a 3D point $\mathbf{x}_\text{lidar} = (x,y,z,1)$ in the rectified LIDAR coordinate will be transformed to $\mathbf{x}_\text{cam}^{(t)} = (x', y', z', 1)$ in the $t^{th}$ rectified camera coordinate with a given matrix $\mathbf{M}_\text{ex}^{(t)}$ called extrinsic parameter.
Next, $\mathbf{x}_\text{cam}^{(t)} = (x', y', z', 1)$ is projected to a point $\mathbf{x}_\text{img}^{(t)}=(u, v, 1)^\top$ in the $t^{th}$ image plane by
\begin{equation}
    \mathbf{x}_\text{img}^{(t)} = \mathbf{M}_\text{in}^{(t)} \mathbf{x}_\text{cam}^{(t)}, \quad \mathbf{M}_\text{in}^{(t)} = \left(
    \begin{array}{c c cc cc c}
        f_u^{(t)} && 0 && c_u && -f_u^{(t)}b_x^{(t)} \\
        0 && f_v^{(t)} && c_v && 0\\
        0 && 0 && 1 && 0 
    \end{array}
    \right)
\end{equation}
Here, $\mathbf{M}_\text{in}^{(t)}$ is the projection matrix for $t^{th}$ camera. 
$(f_u, f_v)$ is the focal length, $(c_u, c_v)$ is the location of optical center and $b_x^{(t)}$ denotes the baseline with respect to reference camera ($0$ for nuScenes).
In summary, a 3D point $\mathbf{x}_\text{lidar}$ can be projected to image point $\mathbf{x}_\text{img}^{(t)}$ by the projection matrix $\mathbf{M}^{(t)} = \mathbf{M}_\text{in}^{(t)}\mathbf{M}_\text{ex}^{(t)}$.
In case of $0<u, v<1$, the point will be projected inside the image, otherwise outside.
If the $ \mathbf{x}_\text{img}$ is inside the image, we say that the image is visible to the corresponding $ \mathbf{x}_\text{lidar} $. 
In our method, imaginary eyes are encouraged to ``look" around their 2D projection point in each image coordinate.
We denote the set of visible images of the $q^{th}$ eyes as $\mathcal{I}_q\subset \mathcal{I}$.

\paragraph{\bf Multi-view multi-scale adaptive attention (\attention)}
\attention~is the core of transferring 2D representation into 3D.
We formulate the learning of 
these imaginary eyes in an adaptive self-attention detection framework \cite{zhu2020deformable}.
This is based on an idea of {\em regarding the eyes as object queries}, denoted as $\mathbf{y}\in\mathbb{R}^{C\times N_\text{eye}}$. 
Let $\mathbf{r}\in \mathbb{R}^{3\times N_\text{eye}}$ be the location of eyes in ego car coordinate.
Formally, each eye (\ie, query) will dynamically choose $N_\text{point}$ feature points at every scale of 2D image representation.
This gives us a total of $N_\text{scale} \times |\mathcal{I}_q |\times N_\text{point}$ feature points.
Our \attention~then chooses the most significant feature points from them and fuse them across multiple scales and views into the desired 3D representation.
The process can be expressed as
\begin{align}
    \notag&\text{\attention}(\mathbf{y}_q, \mathbf{r}_q, \{\{\mathbf{x}_l^{(t)}\}_{l=1}^{N_\text{scale}}\}_{t=1}^{N_\text{view}}) =\\ &\underset{h\in\{N_h\}}{\text{concat}}\mathbf{W}_h\left[
        \sum_{l\in\{N_\text{scale}\}}\sum_{t\in\mathcal{I}_q}\sum_{k\in\{N_\text{point}\}}\mathbf{A}_{hltk}\cdot \mathbf{W}'_h \phi\left(x_l^{(t)}, \mathbf{\mathbf{M}^{(t)}}\mathbf{r} +  \mathbf{\Delta r}_{hvlk}\right) 
    \right]
    \label{equ:mvda}
\end{align}
Vector $\mathbf{y}_q$ is the $q^{th}$ query (eye), $\mathbf{r}_q$ is its position, $N_h$ is the head number, $\mathbf{M}^{(t)}$ is the projection matrix.
$\mathbf{A}$ and $\mathbf{\Delta r}$ are conditioned on $\mathbf{y}_q$ by learnable parameters:
\begin{equation}
\label{equ:adap attention}
    \mathbf{A} = \text{softmax}_{tk}(\mathbf{W}_{q}^{(\mathbf{A})}\mathbf{y}_q)
\end{equation}
 with learnable parameter $\mathbf{W}_{q}^{(\mathbf{A})}\in\mathbb{R}^{N_h\times N_\text{scale}\times |\mathcal{I}_q|\times N_\text{point}\times C}$, and  
\begin{equation}
\label{equ:adap offset}
    \mathbf{\Delta r} = \mathbf{W}_{q}^{(\mathbf{r})}\mathbf{y}_q + \mathbf{b}_q^{(\mathbf{r})}
\end{equation}
with learnable parameter $\mathbf{W}_{q}^{(\mathbf{r})}\in\mathbb{R}^{N_h\times N_\text{scale}\times |\mathcal{I}_q|\times N_\text{point}\times 2\times C}$, and fixed parameter $\mathbf{b}_q^{(\mathbf{r})}\in\mathbb{R}^{N_h\times N_\text{scale}\times |\mathcal{I}_q|\times N_\text{point}\times 2}$.
To avoid these $N_\text{point}$ feature points collapse into one point, $\mathbf{b}_q^{(\mathbf{r})}$ is initialized with $|\mathbf{b}_q^{(\mathbf{r})}[\cdot, \cdot, \cdot, k]| = k$, so that the more $N_\text{point}$, the larger offset of these feature points can be achieved.
Therefore, $N_\text{point}$ can be utilized to control the receptive field.
$\phi(x, r)$ represents access $r^{th}$ feature points from $x$ by index.
For adaptively assigning the significance to $N_\text{scale} \times |\mathcal{I}_q |\times N_\text{point}$ points, the \texttt{Softmax} function
is applied across all the attended feature points, scales, and views: 
\begin{equation}
    \sum\limits_{l\in\{N_\text{scale}\}}\sum\limits_{t\in \mathcal{I}_q }\sum\limits_{k\in\{N_\text{point}\}}\mathbf{A}_{hltk} = 1
\end{equation}

\paragraph{\bf \neck}
Technically, back tracing decoder takes randomly initialized features of imaginary eyes and scales of 2D feature provided by image feature extractor as input, and finally outputs the fine-grained features of imaginary eyes as 3D representation.
\neck~is made up with a stack of 
attention layers adapted from the transformer decoder layers~\cite{misra2021end}.
As shown in Figure~\ref{fig:pipeline}, each layer stacks two self-attention modules and one cross-attention module in order: deformable attention module~\cite{zhu2020deformable}, polar attention and \attention. 
Compared to self-attention, deformable attention is more memory efficient.
On the on 3D representation, we apply a standard self-attention on the eyes of same polar ray for polar attention. 
Also, the feed-forward network (FFN) block is equipped with a depth-wise convolution like \cite{wang2022pvt}.
As illustrated before, \attention\ is responsible for back tracing 2D features into 3D representation.

\subsection{Downstream task head design}
\paragraph{\bf BEV sampling.}
Before the features of imaginary eyes being processed at downstream task, we first grid sample the polarized features into the rectangular ego car coordinate system to match with dataset annotation.
\paragraph{\bf BEV encoder.}
To encode the 3D representation for multiple tasks, we adopt the same BEV encoder module from OFT~\cite{roddick2018orthographic}.
This kind of sub-network is also widely used in the LiDAR-based 3D detector~\cite{yan2018second, lang2019pointpillars}.

\paragraph{\bf Detection head.}
A detection head aims to predict the location, dimension, and orientation of an object in 3D space alongside the object category.
While the previously mentioned stage has generated a dense BEV features, we adopt the popular 3D detection head from CenterPoint~\cite{yin2021center} for our detection task without any modification.

\paragraph{\bf Segmentation head.}
For the BEV segmentation task, we choose a group of progressive up-sampling convolution-based semantic segmentation decoder heads to deal with different elements from the map.
Technically, a $1\times1$ Conv layer, a batch norm layer with ReLU, and a bilinear upsample Conv layer together form one up-sampling module.
The decoder heads for predicting different map elements use the exact BEV features after the BEV encoder.

\vspace{-1em}
\section{Experiments}

\subsection{Setup}
\paragraph{\bf Dataset.} 
We evaluate the proposed model on the nuScenes~\cite{caesar2020nuscenes} dataset, a large-scale autonomous driving dataset with 1000 driving scenes.
Specifically, for multi-camera 3d object detection, it provides streams of images of 6 cameras covering all round from these 1000 scenes, which are then split into 700/150/150 scenes respectively for training, validation, and testing. 
nuScenes also provide informative annotations of the map. 
We choose 5 segmentation tasks: drivable area, pedestrian crossing, walkway, carpark, and divider.

\paragraph{\bf Metrics.}
To evaluate performance, mean Average Precision (mAP)~\cite{everingham2010pascal} and NuScenes Detection Score (NDS)~\cite{caesar2020nuscenes} are reported. 
The mAP is the average of precision over different distance thresholds -- 0.5m, 1m, 2m, 4m, in bird-eye-view. 
NDS is averaged weighted by mAP and True Positive (TP) metrics, where TP metrics is mean of five individual metrics: translation (ATE), velocity (AVE), scale (ASE), orientation (AOE), and attribute (AAE) errors.
The weighted average is calculated by: $\text{NDS} = \frac{1}{10}\left[ 5\text{mAP} + \sum_{\text{mTP}\in\mathbb{TP}} (1-\min(1, \text{mTP}))  \right]$.

The segmentation task uses Intersection over Union(IoU) to assess the performance. As done by the LSS~\cite{philion2020lift}, we create a binary image for each element based on a specific threshold to evaluate with the ground truth image.

\paragraph{\bf Implementation details.}
Following FCOS3D~\cite{wang2021fcos3d} and DETR3D~\cite{wang2022detr3d}, ResNet-101~\cite{he2016deep}, with 3rd and 4th stages equipped with deformable convolutions is applied as our image backbone. 
The following deformable DETR encoder then utilizes multi-scale feature maps from the 2nd, 3rd, and 4th stages of the backbone.
We use eyes of density $80\times 256$ for \name\ and be sampled to rectangular $160\times 160$ grids by BEV sampling.
For the BEV encoder, we use 8 Bottleneck block~\cite{he2016deep} identical to OFT~\cite{roddick2018orthographic}.
In the segmentation task, we set our ground-truth BEV segmentation map of $480 \times 480$ size with 0.2m/pixel resolution.
Therefore, 1 block of the upsampling module with the bilinear upsampling ratio of $3 \times$ is adopted to mitigate lossing details of the screen.
Additionally, Section~\ref{sec:loss} illustrates our loss function.

\paragraph{\bf Training \& testing.}
Our models are trained 24 epoch with \texttt{AdamW}~\cite{loshchilov2017decoupled} of base learning rate $2.5\times10^{-4}$ and weightdecay 0.01. 
Especially, the learning rate of the backbone is 1/10 of the global learning rate and the parameter of batch normalizations of backbone still participate in fine-tuning.
To avoid over-fitting, we apply an early stop at 16 epoch.
Since a total batch size of 48 across six cameras on eight NVIDIA A6000 GPUs is used, we apply synchronized implementation for every batch normalization.
During the training process, we use the input image of $1500\times 900$ resolution with only photometric distortion augmentation.
Random flip, random rotation, and random scaling are applied to the 3D feature.
Our 3D detection head is trained with class-balanced grouping method~\cite{zhu2019class} (but no DS sampling) as default.
After the detection head is well trained, we fix the parameters of \name\ and fine-tune our segmentation head for multi-task.
Random flip, random rotation, and random scaling are applied to the 3D feature when training detection head while no augmentation are used in multi-task training. Details are explained in Appendix~\ref{sec:add_detail}.

\subsection{Comparison with state of the art}
\paragraph{\bf 3D object detection.}

\begin{table*}[t]
  \centering
  \caption{Comparison of different paradigms on the nuScenes \texttt{val} set.
  FCOS3D$\dag$ is trained with 1x learning schedule, depth weight 0.2 and is finetuned on another FCOS3D checkpoint. 
  PGD$\dag$ is trained with 2x learning schedule, depth weight 0.2 on another PGD checkpoint. 
  DETR3D$\dag$ and \name$\dag$ are initialized from the same pretrained FCOS3D checkpoint. 
  \name$\ddag$ is initialized from the pretrained DD3D checkpoint.
  }
\renewcommand{\arraystretch}{1.3}
    \begin{tabular}{l||C{1.3cm}C{1.3cm}C{1.3cm}C{1.3cm}C{1.3cm}||C{1.1cm}C{1.1cm}}
    \hline

    \hline
    \textbf{Methods}        &\textbf{mATE}$\downarrow$   &\textbf{mASE}$\downarrow$   &\textbf{mAOE}$\downarrow$   &\textbf{mAVE}$\downarrow$   &\textbf{mAAE}$\downarrow$ &\textbf{mAP}$\uparrow$  &\textbf{NDS}$\uparrow$  \\
    \hline
    FCOS3D\cite{wang2021fcos3d}  & 0.790 & \textbf{0.261} & 0.499 & 1.286 & \textbf{0.167} & 0.298 & 0.377  \\
    DETR3D\cite{wang2022detr3d}     & 0.860 & 0.278 & \textbf{0.327} & \textbf{0.967} & 0.235 & 0.303 & 0.374  \\
    PGD\cite{wang2022probabilistic} & 0.732 & 0.263 & 0.423 & 1.285 & 0.172& 0.336 & 0.409 \\
    \rowcolor[gray]{.9} 
    \name(Ours)            & \textbf{0.714} & 0.275 & 0.421 & 0.988 & 0.292 & \textbf{0.355} & \textbf{0.409}   \\
    \hline
    
    \hline
    FCOS3D$\dag$\cite{wang2021fcos3d}   & 0.754 & \textbf{0.260} & 0.486 & 1.331 & \textbf{0.158}& 0.321 & 0.395  \\
    DETR3D$\dag$\cite{wang2022detr3d}           & 0.765 & 0.267 & 0.392 & 0.876 & 0.211 & 0.347 & 0.422  \\
    PGD$\dag$\cite{wang2022probabilistic} & 0.667 & 0.264 & 0.435 & 1.276 & 0.177& 0.358 & 0.425  \\
    \rowcolor[gray]{.9} 
    \name(Ours)$\dag$    & \textbf{0.657} & 0.268 & \textbf{0.391} & \textbf{0.850} & 0.206 & \textbf{0.375} & \textbf{0.450}   \\
    \hline
    \hline
    \rowcolor[gray]{.9} 
    \name(Ours)$\ddag$            &\textbf{0.582} & 0.272 & \textbf{0.316} & \textbf{0.683} & 0.202 & \textbf{0.478} & \textbf{0.534}\\
    \hline

    \hline
    \end{tabular}
  \label{tab:nus-det-val}
  \vspace{-1em}
\end{table*}

We compare our model with previous state-of-the-art methods on both nuScenes validation set and test set. 
Following FCOS3D~\cite{wang2021fcos3d} and DETR3D~\cite{wang2022detr3d}, all our experiments are trained using ResNet-101 with deformable convolution as backbone for prototype verification. 
Models without special notification is initialized from a ResNet-101 checkpoint which pre-trained on ImageNet~\cite{deng2009imagenet}.
We also present the result of our model on pre-trained checkpoints from FCOS3D~\cite{wang2021fcos3d} and DD3D~\cite{park2021dd3d}.
In specific, the DD3D~\cite{park2021dd3d} fintunes on extra DDAD15M~\cite{guizilini20203d} dataset.
\textbf{To be noted, the monocular-camera paradigms~\cite{wang2021fcos3d, wang2022probabilistic} and the multi-camera ones can be fairly compared.} 
They all take 6 cameras as input, but the monocular paradigms process these input images independently while the multi-view paradigms process these input images simultaneously.

Table~\ref{tab:nus-det-val} summarizes our multi-camera 3D object detection results on the nuScenes \texttt{validation} set.
The upward arrow means the large the better while the downward one means the small the better.
Our method leads in both \texttt{mAP} and \texttt{NDS}. 
Specially, it achieves the best in \textbf{transition error} (\texttt{mATE}), proving back tracing strategy's ability in localization reasoning.
Just with simple attention, \name\ outperforms localization prediction than the well-designed PGD.

\begin{table*}[htb]
  \centering
  \caption{Comparisons to top-performing works on the nuScenes \texttt{test} set.
  $\ddag$ represents that the method uses external data other than nuScenes 3D box annotations.
  DD3D$\ddag$ uses extra data for depth estimation. 
  DETR3D$\ddag$ and \name$\ddag$ are initialized from the pre-trained DD3D checkpoint.
  }
  \renewcommand{\arraystretch}{1.3}
    \begin{tabular}{l||C{1.3cm}C{1.3cm}C{1.3cm}C{1.3cm}C{1.3cm}|C{1.1cm}C{1.1cm}}
    \hline

    \hline
    \textbf{Methods}        &\textbf{mATE}$\downarrow$   &\textbf{mASE}$\downarrow$   &\textbf{mAOE}$\downarrow$   &\textbf{mAVE}$\downarrow$   &\textbf{mAAE}$\downarrow$ &\textbf{mAP}$\uparrow$  &\textbf{NDS}$\uparrow$  \\
    \hline
    MonoDIS & 0.738 & 0.263 & 0.546 & 1.553 & 0.134 & 0.304 & 0.384\\
    CenterNet~\cite{zhou2019objects} & 0.658 & 0.255 & 0.629 & 1.629 & 0.142 & 0.338  & 0.400  \\
    FCOS3D\cite{wang2021fcos3d} & 0.690 & 0.249 & 0.452 & 1.434& \textbf{0.124} & 0.358 & 0.428  \\
    PGD\cite{wang2022probabilistic} & 0.626 & \textbf{0.245} & \textbf{0.451} & 1.509 & 0.127 & 0.386 & \textbf{0.448}   \\
    \rowcolor[gray]{.9} 
    \name(Ours) & \textbf{0.599} & 0.268 & 0.470 & \textbf{1.169} & 0.172 & \textbf{0.389} & 0.443   \\
    
    \hline

    \hline
    DD3D$\ddag$\cite{park2021dd3d} & 0.572 & \textbf{0.249} & \textbf{ 0.368} & 1.014 & \textbf{0.124} & 0.418 &0.477 \\
    DETR3D$\ddag$\cite{wang2022detr3d}   &0.641 & 0.255 & 0.394 & \textbf{0.845} & 0.133 & 0.412 & \textbf{0.479} \\
    \rowcolor[gray]{.9} 
    \name(Ours)$\ddag$ & \textbf{0.549} & 0.264  & 0.433 & 1.014 & 0.145 & \textbf{0.425} & 0.473\\
    \hline
    
    \hline
    \end{tabular}
  \label{tab:nus-det-test}
  \vspace{-1em}
\end{table*}

\begin{table*}[htb]
  \centering
  \caption{Comparison of BEV semantic segmentation IoU on the nuScenes \texttt{val} set. Multi means wether generate a full surrounded BEV segmentation map from multi-view images. ``-" represents the unprovided result. Single-task version \name\ uses EfficientNet-B0~\cite{tan2019efficientnet} as the image backbone to align with OFT~\cite{roddick2018orthographic} and LSS~\cite{philion2020lift}.
  Multi-task version \name$ \P $ means we only train the segmentation head with the pretrained detection model frozen.}
    \begin{tabular}{lc||cccccc}
    \hline
    
    \hline
    \textbf{Method}      & \textbf{multi?}     &\textbf{Drivable}  &\textbf{Crossing}  &\textbf{Walkway}   &\textbf{Carpark} &\textbf{Divider}\\
    \hline
    VED~\cite{Lu2019icra-ral}               &\XSolidBrush           &54.7 &12.0 &20.7 &13.5  &- \\
    VPN~\cite{pan2019crossview}             &\XSolidBrush        &58.0 &27.3 &29.4 &12.3  &- \\
    PON~\cite{roddick2020predicting}        &\XSolidBrush        &60.4 &28.0 &31.0 &18.4  &- \\
    OFT~\cite{roddick2018orthographic}      &\XSolidBrush        &62.4 &30.9 &34.5 &23.5  &-  \\
    LSF~\cite{dwivedi2021bird}              &\XSolidBrush        &61.1 &33.5 &37.8 &25.4  &- \\    
    Image2Map~\cite{saha2021translating}    &\XSolidBrush        &74.5 &\textbf{36.6} &35.9 &31.3  &- \\
    \hline
    OFT~\cite{roddick2018orthographic}      &\checkmark       &71.7 &- &- &- &18.0     \\
    LSS~\cite{philion2020lift}              &\checkmark       &72.9 &- &- &- &20.0     \\                 
    \rowcolor[gray]{.9} 
    \name(Ours)                             &\checkmark         &\textbf{79.6} &\textbf{48.3} &\textbf{52.0} & \textbf{50.3} &\textbf{47.5}     \\
    \rowcolor[gray]{.9} 
    \hline
    \hline
    \name(Ours) $ \P $                      &\checkmark              &\textbf{74.6} &33.0 &\textbf{42.6} & \textbf{44.1} &\textbf{36.6}     \\
    
    \hline
    
    \hline
    \end{tabular}
  \label{tab:nus-seg_val}
  \vspace{-1em}
\end{table*}

Table~\ref{tab:nus-det-test} shows our results on the nuScenes \texttt{test} set. 
The training sets are the same as the validation set. 
The least \textbf{transition error} (\texttt{mATE}) also reflects the overwhelming localization power of back tracing mechanism.
We achieve the best \texttt{mAP} but the \texttt{NDS} is hindered by the attribute error.
We have to say that the 2D representation has a manifest advantage on classification over 3D representation.

\paragraph{\bf BEV segmentation.}

Table~\ref{tab:nus-seg_val} summarizes our BEV segmentation results on nuScenes \texttt{validation} set.
We achieve the best performance in all tasks except the pedestrian crossing, but its overwhelming advantage in other tasks still proves its success.
We also present single-task version using ImageNet pre-trained Efficient-B0\cite{tan2019efficientnet} as our image backbone to make a fair comparison with OFT\cite{roddick2018orthographic} and LSS~\cite{philion2020lift}.
In terms of the single-task version, \name\ leads the board with huge superiority.

\subsection{Qualitative results}

\begin{figure}[htb]
    \centering
    \includegraphics[width=0.95\linewidth]{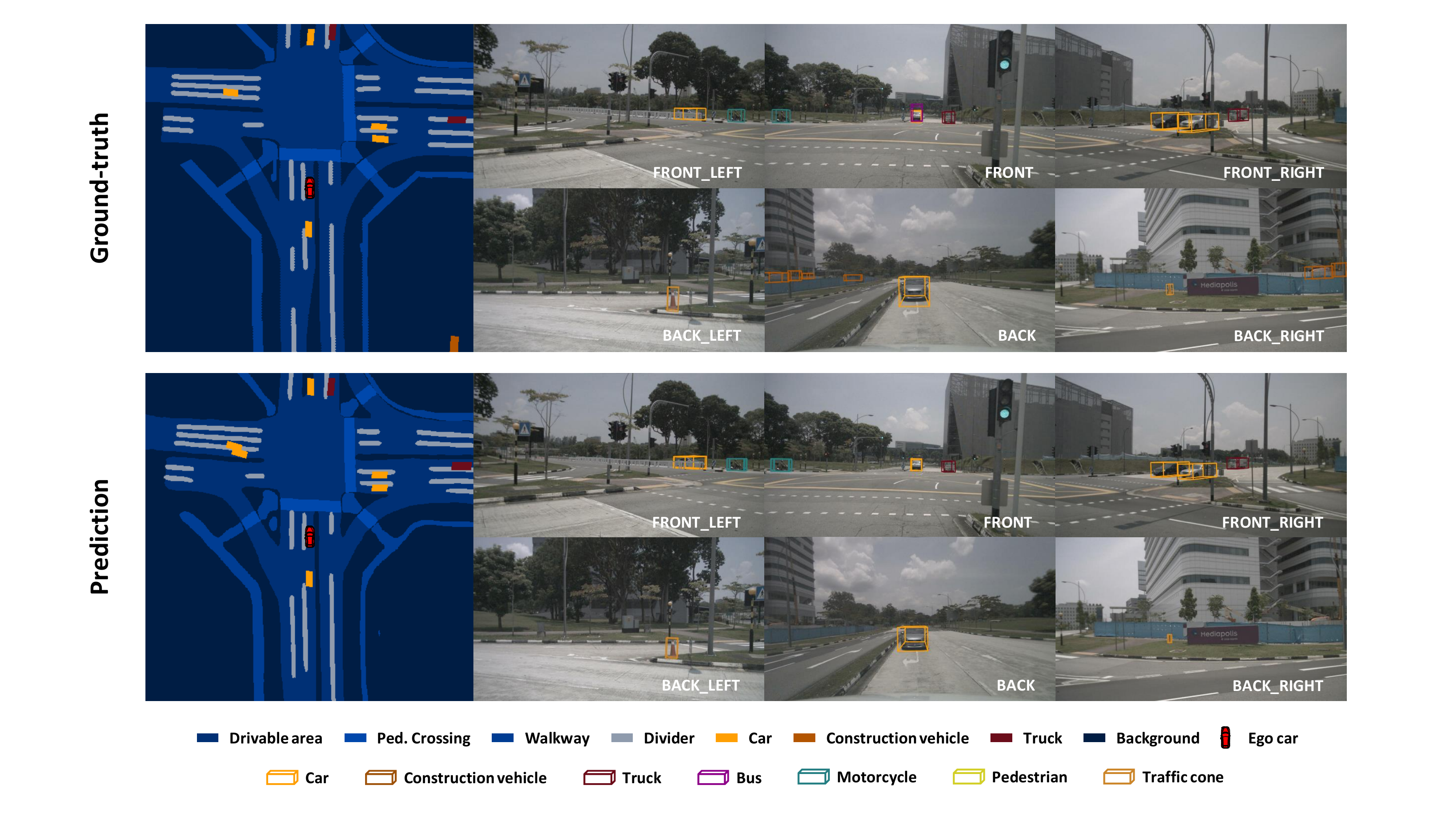}
    \caption{Qualitative results on nuScenes dataset. 
    {\em Left} is the BEV segmentation map with projected {\em vehicle} 3D bounding boxes result from detection head.
    {\em Right} are in image perspective with prediction results.
    Different colors stand for different categories.
    }
    \label{fig:visualization}
    \vspace{-2em}
\end{figure}

We present our visualization in Figure~\ref{fig:visualization}. 
The structural similarity to the ground-truth highlights the superiority of our \name{}, which simultaneously generates dynamic object detection and static semantic segmentation results from the 3D representation.
In specific, we project all bounding boxes of class {\em vehicle} in nuScenes from the detection head onto the generated BEV segmentation map for a clear comparison.
As can be seen from the perspective of images, our detection results demonstrate appealing localization ability even in distance situations.
Section~\ref{sec:add_vis} provides more qualitative results

\subsection{Ablation studies}
In this section, we will figure out how the performance is established and prove the effectiveness of our innovation.

\paragraph{\bf Image backbone.} 
\begin{table*}[htb]
  \centering
  \caption{Comparison of different image feature extractors. 
  $\dag$ means the image feature extractor is initialized from a FCOS3D checkpoint. 
  $\ddag$ means the image feature extractor is initialized from a DD3D checkpoint. 
  }
\renewcommand{\arraystretch}{1.2}
    \begin{tabular}{l||C{1.3cm}C{1.3cm}C{1.3cm}C{1.3cm}C{1.3cm}|C{1.1cm}C{1.1cm}}
    \hline

    \hline
    \textbf{Backbone}        &\textbf{mATE}$\downarrow$   &\textbf{mASE}$\downarrow$   &\textbf{mAOE}$\downarrow$   &\textbf{mAVE}$\downarrow$   &\textbf{mAAE}$\downarrow$ &\textbf{mAP}$\uparrow$  &\textbf{NDS}$\uparrow$  \\
    \hline
    ResNet50       &0.706 & 0.281 & 0.663 & 0.964 & 0.249 & 0.332 & 0.380  \\
    ResNet101   & 0.714 & 0.275 & 0.421 & 0.988 & 0.292 & 0.355 & 0.409   \\
    ResNet101 $\dag$ & 0.657 & \textbf{0.268} & 0.391 & 0.850 & 0.206 & 0.375 & 0.450\\
    VoveNet$\ddag$   &\textbf{0.582} & 0.272 & \textbf{0.316} & \textbf{0.683} & \textbf{0.202} & \textbf{0.478} & \textbf{0.534}\\
    
    \hline

    \hline
    \end{tabular}
  \label{tab:backbone}
  \vspace{-1em}
\end{table*}

We first provide results with different image feature extractors in Table~\ref{tab:backbone}. 
It presents that the learning of 3D representation relies highly on 2D representation.

\paragraph{\bf Back tracing mechanism}
\begin{table*}[htb]
  \centering
  \caption{Comparisons of detection performance in non-overlap region and overlap region. 
  FCOS3D is trained with 1x learning schedule, depth weight 0.2 and is finetuned on another FCOS3D checkpoint.  
  PGD is trained with 2x learning schedule, depth weight 0.2 and is finetuned on another PGD checkpoint. 
  DETR3D and \name are initialized from a same pretrained FCOS3D checkpoint. 
  }
\renewcommand{\arraystretch}{1.2}
    \begin{tabular}{lC{1.3cm}||C{1.2cm}C{1.2cm}C{1.2cm}C{1.2cm}C{1.2cm}|C{1cm}C{1cm}}
    \hline

    \hline
    \textbf{Methods}       & \textbf{overlap?} &\textbf{mATE}$\downarrow$   &\textbf{mASE}$\downarrow$   &\textbf{mAOE}$\downarrow$   &\textbf{mAVE}$\downarrow$   &\textbf{mAAE}$\downarrow$ &\textbf{mAP}$\uparrow$  &\textbf{NDS}$\uparrow$  \\
    \hline
    FCOS3D\cite{wang2021fcos3d}& \XSolidBrush & 0.747  & \textbf{0.260} & 0.487 & 1.351 & \textbf{0.156} & 0.320 & 0.395    \\
    PGD\cite{wang2022probabilistic}& \XSolidBrush & 0.658 & 0.263 & 0.425 & 1.290 & 0.178 &  0.357 & 0.426   \\
    DETR3D\cite{wang2022detr3d}     & \XSolidBrush      & 0.769 & 0.267 & \textbf{0.390} & 0.893 & 0.215 & 0.343 & 0.419  \\
    \rowcolor[gray]{.9} 
    \name(Ours)   & \XSolidBrush        & \textbf{0.655} & 0.267 & 0.395 & \textbf{0.854} & 0.208 & \textbf{0.371} & \textbf{0.448}   \\
    
    \hline

    \hline
    
    FCOS3D\cite{wang2021fcos3d} & \checkmark & 0.816 & 0.272 & 0.571 & 1.084 & \textbf{0.173} & 0.229 & 0.329    \\
    PGD\cite{wang2022probabilistic}& \checkmark & 0.768 & 0.274 & 0.495 & 1.090 & 0.186 &  0.255 & 0.354   \\
    DETR3D\cite{wang2022detr3d}  & \checkmark        & 0.807 & 0.273 & 0.453 & \textbf{0.788} & 0.184 & 0.268 & 0.384  \\
    \rowcolor[gray]{.9} 
    \name(Ours)      &\checkmark      & \textbf{0.671} & \textbf{0.268} & \textbf{0.347} & 0.797 & 0.212 & \textbf{0.298} & \textbf{0.420}   \\
    \hline
    
    \hline
    \end{tabular}
  \label{tab:overlap}
  \vspace{-1em}
\end{table*}

\begin{table*}[htb]
  \centering
  \caption{Ablation on the effectiveness of adaptive attention mechanism. 
  }
\renewcommand{\arraystretch}{1.2}
    \begin{tabular}{C{1.7cm}||C{1.2cm}C{1.2cm}C{1.2cm}C{1.2cm}C{1.2cm}|C{1.0cm}C{1.0cm}}
    \hline

    \hline
   \textbf{ adaptive?} &\textbf{mATE}$\downarrow$  &\textbf{mASE}$\downarrow$  &\textbf{mAOE}$\downarrow$   &\textbf{mAVE}$\downarrow$ &\textbf{mAAE}$\downarrow$
    &\textbf{mAP}$\uparrow$
    &\textbf{NDS}$\uparrow$  \\
    \hline
     \XSolidBrush & 0.688 & 0.272 & 0.403 & \textbf{0.835} & 0.217 & 0.365 & 0.441\\
     \Checkmark& \textbf{0.657} & \textbf{0.268} & \textbf{0.391} & 0.850 & \textbf{0.206} & \textbf{0.375} & \textbf{0.450}\\
    
    \hline

    \hline
    \end{tabular}
    \vspace{-1em}
  \label{tab:adaptive}
\end{table*}

Here, we prove that the back tracing mechanism posses superiority in localization.
To eliminate interference of multi-view mechanism, in the top part of Table~\ref{tab:overlap}, hence, we validate the 3D detection result only at the non-overlap region where only a monocular camera is used.
We show advantage in overall \texttt{mAP}, \texttt{NDS} metrics at monocular region. 
Specially, the lowest transition error \texttt{mATE} proves the best localization reasoning of \name{} .
\paragraph{\bf Multi-view mechanism}
We will prove the superiority of the multi-view mechanism over the former monocular ones.
In the bottom part of Table~\ref{tab:overlap}, we validate the 3D detection result at region where only multiple cameras are used.
We find that multi-view methods DETR3D and our \name\ outperform mono-view methods FCOS3D~\cite{wang2021fcos3d} and PGD~\cite{wang2022probabilistic} remarkably in all metrics.
Additionally, \name\ achieves overwhelming performance over the other methods in both \texttt{mAP} and \texttt{NDS}.

\paragraph{\bf Adaptive attention mechanism}

We state in the method section that the adaptive attention mechanism can approximate missing height information.
All the other conditions remaining the same, we switch off the adaptive attention module by fixing learnable parameter $\mathbf{W}_{q}^{(\mathbf{r})}\mathbf{y}_q$ in Eq.~\eqref{equ:adap offset} to prove its effectiveness.
The results shown in Table~\ref{tab:adaptive} prove our statement. 

\paragraph{\bf Polarized grid of imaginary eyes}
\begin{table*}[htb]
  \centering
  \caption{Ablations on polarized grid of imaginary eyes.}
\renewcommand{\arraystretch}{1.2}
    \begin{tabular}{C{1.3cm}C{1.7cm}||C{1.2cm}C{1.2cm}C{1.2cm}C{1.2cm}C{1.2cm}|C{1.0cm}C{1.0cm}}
    \hline

    \hline
    \textbf{polarized} & \textbf{polar} & 
    \multirow{2}{*}{\textbf{mATE}$\downarrow$} & \multirow{2}{*}{\textbf{mASE}$\downarrow$} &\multirow{2}{*}{\textbf{mAOE}$\downarrow$} & \multirow{2}{*}{\textbf{mAVE}$\downarrow$} & \multirow{2}{*}{\textbf{mAAE}$\downarrow$} & \multirow{2}{*}{\textbf{mAP}$\uparrow$} & \multirow{2}{*}{\textbf{NDS}$\uparrow$} \\
    \textbf{grid?} & \textbf{attention?}&&&&&&&\\
    \hline
    \XSolidBrush &\XSolidBrush        &0.673 &0.271 &0.397 & 0.901 & 0.211 & 0.365 & 0.437\\
    \Checkmark  &\XSolidBrush         & \textbf{0.656} & 0.271 & 0.397 & 0.881 & \textbf{0.206} & 0.362 & 0.440\\
    \Checkmark  &\Checkmark         & 0.657 & \textbf{0.268} & \textbf{0.391} & \textbf{0.850} & \textbf{0.206} & \textbf{0.375} & \textbf{0.450}\\
    \hline

    \hline
    \end{tabular}
  \label{tab:polarize}
\end{table*}

We apply a polarized grid of ``imaginary eyes" rather than a rectangular grid.
Here, we compare these two settings in Table~\ref{tab:polarize}.
The first line and the second line compare the grid type of eyes without polar attention.
Although the polarized grid doesn't show manifest superiority over the rectangular grid, the polarized grid achieves a remarkable advantage in localization prediction (\texttt{mATE}). 
Finally, with the help of polar attention on the eyes of each polar ray, the model achieves the best. 

\begin{table}[tb]
    \centering
    \caption{(a) Results with different density of imaginary eyes. (b) Results with different height above the ground of imaginary eyes in meter.}

    \subfloat{
        \begin{tabular}[b]{C{1cm}||C{1.4cm}C{1.4cm}C{1.4cm}C{1.4cm}}
        \hline
        
        \hline
        \#\textbf{eyes} & $\mathbf{96\times256}$ & $\mathbf{64\times256}$ & $\mathbf{80\times224}$ & $\mathbf{80\times256}$ \\
        \hline
        mAP $\uparrow$ & 0.372 & 0.366 & 0.372  & \textbf{0.375} \\
        
        NDS $\uparrow$ & 0.447 & 0.444 & 0.445  &\textbf{ 0.450} \\
        \hline
        
        \hline
        \end{tabular}
    }
    \subfloat{
        \begin{tabular}[b]{C{1cm}||C{1cm}C{1cm}C{1cm}}
        \hline
        
        \hline
        \textbf{height} & \textbf{0m} &\textbf{0.4m} & \textbf{0.8m} \\
        \hline
        mAP $\uparrow$ & 0.375 & 0.375 & 0.375   \\
        
        NDS $\uparrow$ & 0.445 & 0.437 & \textbf{0.450}  \\
        \hline
        
        \hline
        \end{tabular}
    }
    \label{tab:dense_height}
    \vspace{-1em}
\end{table}
\begin{table}[htb]
    \centering
    \vspace{-1em}
    \caption{Ablation on each component of Ego3RT. For the baseline, we set $N_{point}=1$ (w/o ``looking around"), eliminate 3D feature augmentation, adaptive attention mechanism (w/o ``adaptive looking") and polarization (including both polarized grid and polar attention).}
    \vspace{-1em}
    \begin{tabular}{l ||cc}
    \hline
    
    \hline
    \textbf{Components} &\textbf{mAP}$\uparrow$  & \textbf{NDS}$\uparrow$ \\
    \hline
    baseline & 0.345 & 0.421 \\
    \quad+3D feature augmentation & 0.353 & 0.427 \\
    \quad+$N_{point}=3$ & 0.360 & 0.433\\
    \quad+adaptive attention & 0.365 & 0.437 \\
    \quad+polarization & 0.375 & 0.450\\

    \hline
    
    \hline
    \end{tabular}
    \label{tab:improvement}
    \vspace{-1em}
\end{table}
\paragraph{\bf Density of imaginary eyes} 
In this section, we compare imaginary eyes of different densities.
Lower density will lead to coarser feature maps while the higher density imposes a burden on optimization, so a balance should be achieved.
Table~\ref{tab:dense_height}(a) shows that our final choice $80\times 256$ is the optimal choice.

\paragraph{\bf Height of imaginary eyes}
Since we use the imaginary eyes occupied at a pre-given location, the height is a hyper-parameter. 
We compares different choice of height in Table~\ref{tab:dense_height}(b). It shows that no significant difference exists among the different choices of height.

\paragraph{\bf Which leads to improvement}
To further clarify, we summarize the effect of each component in Table~\ref{tab:improvement}, including \changed{3D feature augmentation}, the choice of $N_{point}$, adaptive attention mechanism, polarized grid and polar attention.
Importantly, each component of our Ego3RT yields good gain.
\begin{table*}[h]
  \centering
  \caption{Comparison of the efficiency and the performance of different configurations of Ego3RT and the other methods. 
  ``FPS" is a metric for efficiency standing for frames per second. 
  ``Resolution" represents input image shape.
  ``FFN" represents the channel expansion dimension of FFN in Back tracing decoder.
  ``Blocks" notes the number of blocks in BEV encoder.
  ``$\star$" means we test the speed at $1600\times 900$.
  }
\renewcommand{\arraystretch}{1.2}
    \begin{tabular}{l||cccc|C{1.1cm}|C{1.1cm}C{1.1cm}}
    \hline

    \hline
    \textbf{Methods} & \textbf{Resolution} & \textbf{Eyes density} & \textbf{FFN} & \textbf{\#Blocks} & \textbf{FPS}$\uparrow$ &\textbf{mAP}$\uparrow$  &\textbf{NDS}$\uparrow$  \\
    \hline
    FCOS3D\cite{wang2021fcos3d} & $1600\times900$ & - & - & - & 2.0 &0.321 & 0.395\\
    PGD~\cite{wang2022probabilistic} & $1600\times900$ & -& -  & -& 1.5 & 0.358 & 0.425 \\
    DETR3D~\cite{wang2022detr3d} & $1600\times900$ & - & -  & -& \textbf{3.0}& 0.347 & 0.422\\
    \hline
    Ego3RT(Ours) & $1600\times900^\star$ & $80\times 256$ & 1024 & 8 &1.7 & \textbf{0.375} & \textbf{0.450}\\
    Ego3RT(Ours) & $1280\times768$ & $72\times 192$ & 1024& 8 &2.3 & 0.372 & 0.438\\
    Ego3RT(Ours) & $1280\times768$ & $64\times 128$ & 256 & 2 &\textbf{3.0} & 0.355 & 0.423\\
    \hline

    \hline
    \end{tabular}
  \label{tab:fps}
  \vspace{-1em}
\end{table*}

\paragraph{\bf Efficiency of Ego3RT}
We compare the efficiency and performance of different Ego3RT configurations and the other methods in Table~\ref{tab:fps}.
Our Ego3RT (of main configuration) achieves the best mAP and 3rd FPS, and Ego3RT with smaller input image size and imaginary eyes density barely looses its performance while achieves a better efficiency.
When Ego3RT further reduces its eyes density, FFN channel expansion dimension and BEV encoder blocks, it reaches 
the best trade-off between accuracy and efficiency.
\section{Conclusion}
In this work, we have presented \name, a novel end-to-end architecture
for ego 3D representation learning given multiple unconstrained camera views.
In the absence of depth or 3D supervision, 
it can learn rich and semantic 3D representation with multi-view images 
efficiently in the ego car coordinate frame.
This is realized by drawing an analog from the ray tracing concept,
where we create a polarized grid of learnable ``imaginary eyes" as the ego 3D representation and formulate the learning with the adaptive attention mechanism subject to the 3D-to-2D projection.
It is easy to implement and able to support multiple different tasks.
Extensive experiments validate the superiority of our \name~in comparison to state-of-the-art alternatives in terms of both accuracy and versatility.

\section*{Acknowledgments}
This work was supported in part by 
National Natural Science Foundation of China (Grant No. 6210020439),
Lingang Laboratory (Grant No. LG-QS-202202-07),
Natural Science Foundation of Shanghai (Grant No. 22ZR1407500).

% \clearpage

\bibliographystyle{splncs04}
\bibliography{egbib}
\appendix

\section{Appendix}

\subsection{Objective functions}
\label{sec:loss}
There are two training objectives for our model, including the loss $\mathcal{L}_{det}$ for object detection, and the loss $\mathcal{L}_{seg}$ for map elements segmentation:
\begin{equation}
    \mathcal{L}_{} = \mathcal{L}_{cls} + \mathcal{L}_{seg} .
\end{equation}

\paragraph{\bf Detection}
To handle the severe class imbalance with the nuScenes dataset, following CBGS~\cite{zhu2019class} we group the similar classes into the same sub-task head.
We use the focal loss for classification to alleviate the sample imbalance during our training, and simply adopt $L1$ loss to regress the normalized box parameters.

The classification loss for a specific sub-task $\mathcal{L}_{cls}^{t}$ is formulated as follows: 
\begin{equation}
 \begin{array}{r}
  \mathcal{L}_{cls}^{t}= - \frac{1}{N} \sum_{y_{cls} \in Y_{cls}}
  \begin{cases}
        (1 - \hat{y}_{cls})^{\alpha} 
        \log(\hat{y}_{cls}) & \!\text{if}\ y_{cls}=1\vspace{2mm}\\
        \begin{array}{c}
        (1-y_{cls})^{\beta} 
        (\hat{y}_{cls})^{\alpha} \log(1-\hat{y}_{cls})
        \end{array}
        & \!\text{otherwise}
    \end{cases}
    ,
 \end{array}
\label{loss_cls}
\end{equation}
where $Y_{cls}$ and N represents the set of pixels on the heatmap and the number of objects in $t$-th group, respectively. 
$\hat{y}_{cls}$ is the predicted classification probability and ${y}_{cls}$ is the ground-truth. 
$\alpha$ and $\gamma$ are the parameters of the focal loss\cite{lin2017focal}. 

The 3D bounding box regression loss for a specific sub-task $\mathcal{L}_{box}^{t}$ could be formulated as:
\begin{equation}
    \mathcal{L}_{box}^{t} =  \sum_{res \in \mathcal{R}} \mathcal{L}_{L1}(\widehat{\Delta_{res}}, \Delta_{res}),
\label{loss_box}
\end{equation}
where $\widehat{\Delta_{res}}$ is the predicted residual for the candidate center and $\Delta_{res}$ is the target ground-truth. $\mathcal{R}$ is the set of a box parameters, where $x,y$ are the refinement for the location, $z$ stands for the height, $l,h,w$ are the 3D bounding box size, $\sin{\theta}$ and $\cos{\theta}$ are the rotation at yaw angel, $v_x$, $v_y$ represent the velocities of the object.

Therefore, the overall detection loss $\mathcal{L}_{det}$ is formulated:
\begin{equation}
    \mathcal{L}_{det} = \sum_{t \in \mathcal{T}_{det} } ( \lambda_{cls} \mathcal{L}_{cls}^{t} +  \lambda_{box} \mathcal{L}_{box}^{t}),
\label{loss_det}
\end{equation}
where $\mathcal{T}_{det}$ stands for the set of sub-task groups, $\lambda_{cls}$ and $\lambda_{box}$ represent the loss weights for classification and box regression.

\paragraph{\bf Segmentation}
We use 5 different segmentation heads for the static elements in the BEV map, and the pixel-wise binary cross-entropy loss $\mathcal{L}_{seg}^{t}$ for $t$-th sub-task.
The overall segmentation loss $\mathcal{L}_{seg}$ is computed as follows:
\begin{equation}
    \mathcal{L}_{seg} = \sum_{t \in \mathcal{T}_{seg} } \lambda_{seg}^{t} \mathcal{L}_{seg}^{t} ,
\label{loss_seg}
\end{equation}
where $\mathcal{T}_{seg}$ represents the set of elements in the BEV map, $\lambda_{seg}^{t}$ is the loss weight of the element.

\subsection{Additional training \& testing details}
\label{sec:add_detail}
When training detection head, rotation degree is uniformly selected from $-22.5^\circ\sim 22.5^\circ$ to perform random rotation. 
Then, scale ration is uniformly selected from $0.95\sim 1.05$ to perform random scale.
Next, horizontal flip and vertical flip are uniformly selected to perform random flip.
All the augmentations are applied in 50\% probability.
Critically, to make fair comparison, \textbf{no 3D feature augmentation} are used in BEV segmentation training and no test time augmentation in all tasks.
The effectiveness of 3D feature augmentation is shown in Table~\ref{tab:improvement} that the augmentation brings 2\% improvement to mAP. 
However, \cite{huang2021bevdet} states that the 3D feature augmentation brings at least 20\% improvement to a depth-based method.
It reveals that the 3D representation generated by the depth-based method~\cite{philion2020lift} (3D-2D relationship is based on depth estimation with only a few convolution layers) suffers severe over-fitting in detection tasks without 3D augmentation.
In contrast, 3D-2D relationship of our Ego3RT shows a robustness with the help of adaptive attention mechanism.

\subsection{Additional qualitative results}
\label{sec:add_vis}
\paragraph{\bf Visualization with video}
On our page, we simultaneously generate visualization of dynamic object detection and static semantic segmentation results from the 3D representation.
In specific, we project all bounding boxes of class {\em vehicle} in nuScenes from the detection head onto the generated BEV segmentation map for a clear comparison.

\begin{figure}[htb]
    \centering
    \subfloat{
    \includegraphics[width=\linewidth]{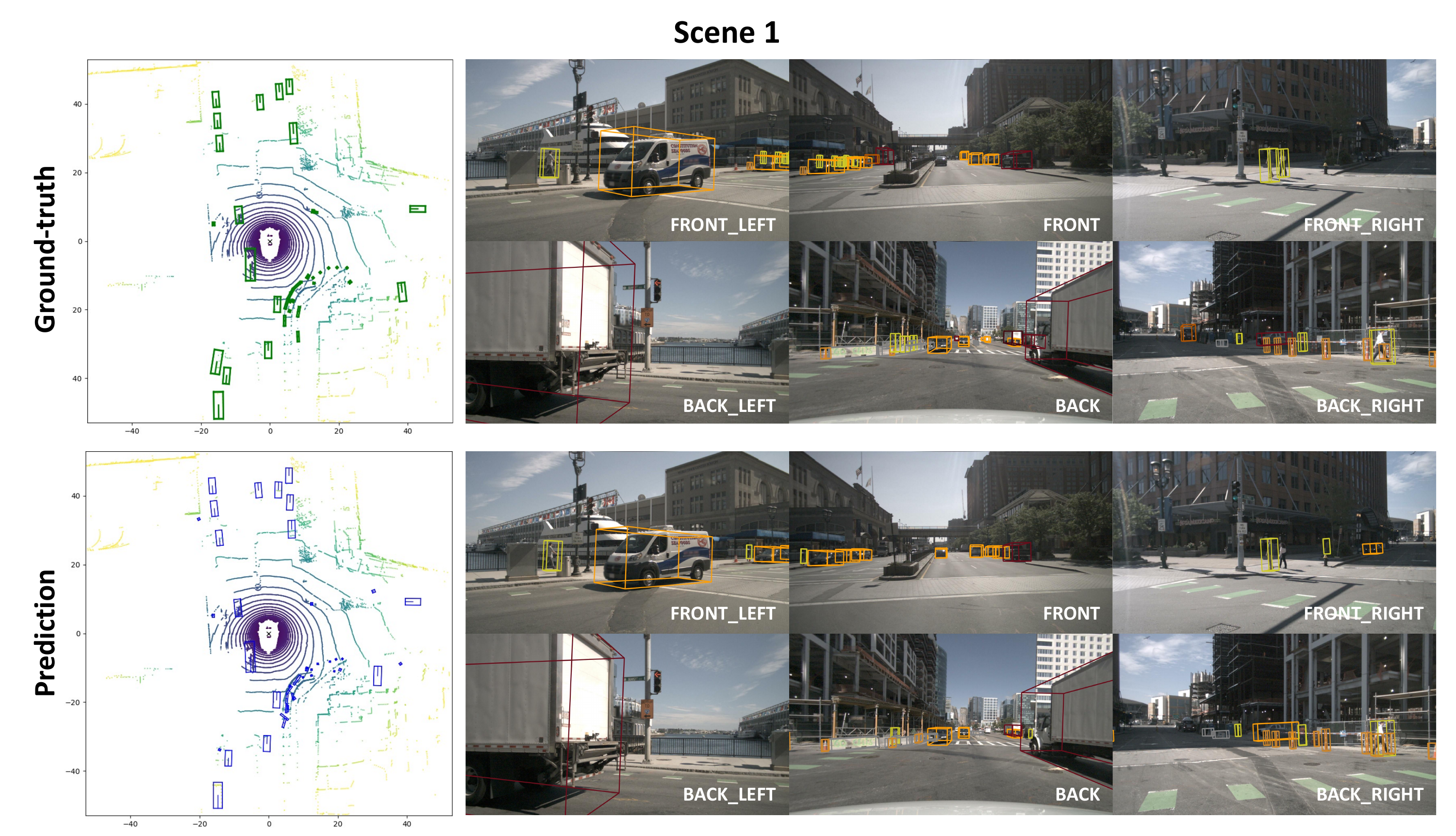}
    }
    \quad
    \subfloat{
    \includegraphics[width=\linewidth]{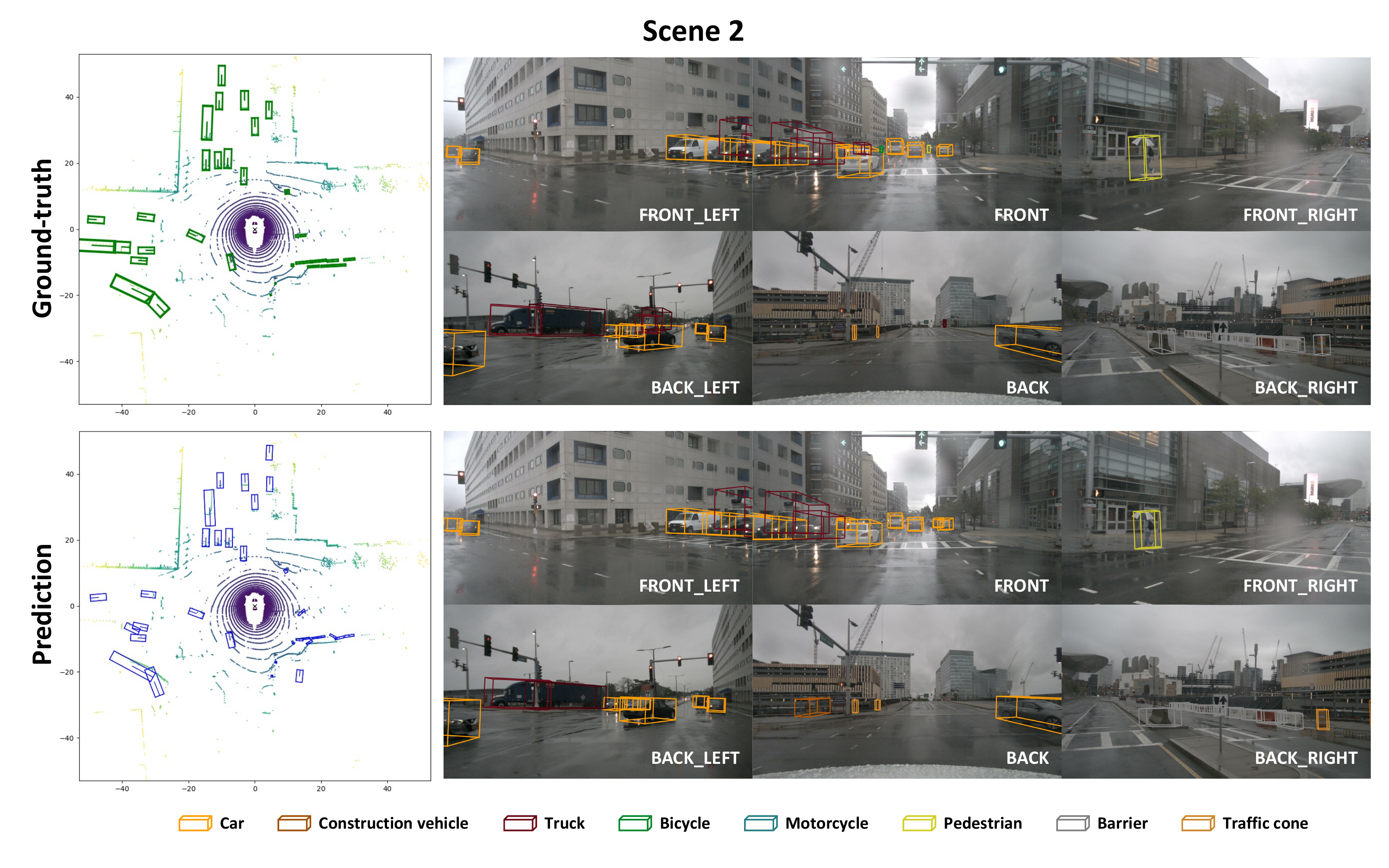}
    }
    \caption{Qualitative results on nuScenes dataset. 
    Two scenes with both ground-truth and prediction are shown.
    {\em Left} are bird's-eye-view visualizations of object detection results.
    {\em Right} are in image perspective with prediction results.
    Different colors stand for different categories.
    }
    \label{fig:append_vis_det}
\end{figure}

\paragraph{\bf Visualization of object detection results}
Figure~\ref{fig:append_vis_det} presents visualization of object detection results of two scenes in nuScenes \texttt{val} set.
We have the following observations.
(i) \name{} yields precise localization regarding to the bird's-eye-view visualization,
even for the objects at long distance.
(ii) \name{} can still work well in rainy whether shown in the second scene, proving its robustness to the whether condition.
(iii) There are some miss-labeling in this dataset.
For example, traffic cone in the \texttt{BACK\_RIGHT} image of second scene is mis-labeled as barrier, but \name{} correctly labels it as traffic cone.

\begin{figure}[htb]
    \centering
    \includegraphics[width=\linewidth]{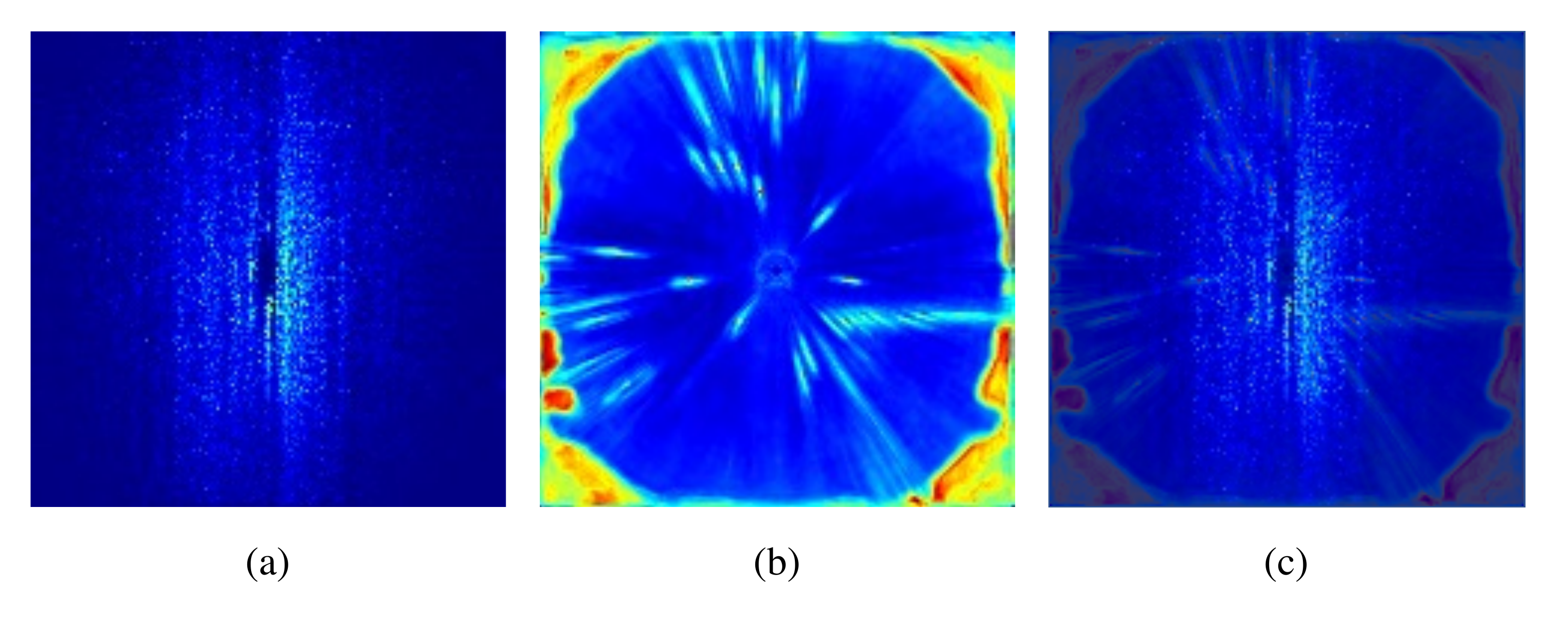}
    \caption{
    (a) Distribution (heat map) of object localization  in bird's-eye-view on nuScenes dataset. 
    Colors closer to \texttt{Red} represent higher frequency while colors closer to \texttt{Blue} represent lower frequency. (b) 3D representation generated by \name{}. Colors closer to \texttt{Red} represent higher response while colors closer to \texttt{Blue} represent lower response. 
    (c) Distribution of object localization with the 3D representation, in the same coordinate as bird's-eye-view.
    }
    \label{fig:vis_bev_gtdist}
\end{figure}

\paragraph{\bf Visualization of 3D representation}
\begin{figure}[htb]
    \centering
    \includegraphics[width=\linewidth]{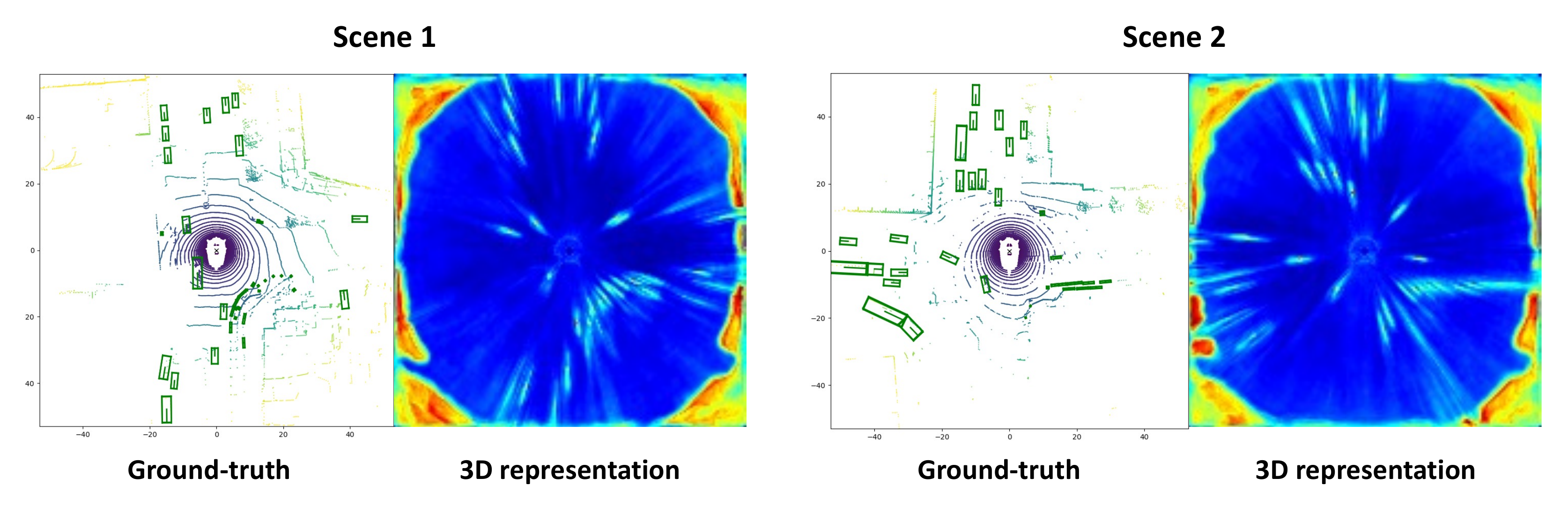}
    \caption{Visualization results of 2 scenes' 3D representations given by \name{}. For each scene, {\em left} is the ground-truth of objects in bird's-eye view, while {\em right} is the visualization of 3D representation. Colors closer to \texttt{Red} represent higher response while colors closer to \texttt{Blue} represent lower response. 
    }
    \label{fig:vis_bev}
\end{figure}
We provide the visualization of \name{}'s learned 3D representation of the same scenes shown in the last section in Figure~\ref{fig:vis_bev}.
The 3D representations predicted by \name{} are simply taken average on the channel to visualize.
There is clear activation in the 3D representation wherever there is an object.
The visualization demonstrates that \name{} actually learns 3D dense representation.
\paragraph{\bf Objects' localization distribution}
There is an interesting observation that the outer part of 3D representation has different pattern in comparison with the inner part.
At the beginning, we considered it was caused by the error in codes, but this different pattern remained even after a careful inspection.
It is not until we visualized the objects' localization distribution of nuScenes that the answer was uncovered.
Objects in nuScenes dataset appear more frequently at the center than the surrounding area.
As is shown in Figure~\ref{fig:vis_bev_gtdist}(c), the boundary of 3D representation's inner part well matches that of the objects' localization distribution. 
Therefore, \name{} reveals some data distribution while reasoning the 3D representation.
\end{document}